\documentclass{article} 

\usepackage{iclr2016_conference,times}
\usepackage{times,hyperref,amsfonts,amsmath,graphicx,wrapfig,enumitem,amssymb,diagbox}
\usepackage{multirow}
\usepackage{caption}

\usepackage{epigraph}
\definecolor{LimeGreen}{rgb}{0.1,0.9,0.1}
\definecolor{Maroon}{rgb}{0.9,0.1,0.1}
\definecolor{bananayellow}{rgb}{1.0, 0.88, 0.21}
\definecolor{cadmiumyellow}{rgb}{1.0, 0.96, 0.0}
\newcommand{\no}{ $\color{Maroon}\times$ }
\newcommand{\yes}{ $\color{LimeGreen}\checkmark$  }

\title{Reinforcement Learning\\Neural Turing Machines - Revised}

\author{Wojciech Zaremba$^{1,2}$ \\
New York University\\
Facebook AI Research\\
\texttt{woj.zaremba@gmail.com} \\
\And
Ilya Sutskever$^2$ \\
Google Brain \\
\texttt{ilyasu@google.com} \\
}

\begin{document}

\maketitle

\vspace{-0.5cm}

\begin{abstract}
The Neural Turing Machine (NTM) is more expressive than all
previously considered models because of its external memory. 
It can be viewed as a broader effort to use abstract external
\textit{Interfaces} and to learn a parametric model that interacts
with them.

The capabilities of a model can be extended by providing it with
proper Interfaces that interact with the world.  These external
Interfaces include memory, a database, a search engine, or a piece of
software such as a theorem verifier.  Some of these Interfaces are
provided by the developers of the model.  However, many important
existing Interfaces, such as databases and search engines, are
discrete.

We examine feasibility of learning models to interact with discrete
Interfaces. We investigate the following discrete Interfaces: a memory
Tape, an input Tape, and an output Tape.  We use a Reinforcement
Learning algorithm to train a neural network that interacts with such
Interfaces to solve simple algorithmic tasks.  Our Interfaces are
expressive enough to make our model Turing complete.
\end{abstract}

\footnotetext[1]{Work done while the author was at Google.}
\footnotetext[2]{Both authors contributed equally to this work.}


\vspace{-0.5cm}

\section{Introduction}

\cite{ntm}'s Neural Turing Machine (NTM) is model that learns to
interact with an external memory that is differentiable and
continuous.  An external memory extends the capabilities of the NTM,
allowing it to solve tasks that were previously unsolvable by
conventional machine learning methods.  This is the source of the
NTM's expressive power.  In general, it appears that ML models become
significantly more powerful if they are able to learn to interact with
external interfaces.

There exist a vast number of Interfaces that could be used with our
models. For example, the Google search engine is an example of such
Interface. The search engine consumes queries (which are actions),
and outputs search results. However, the search engine is not
differentiable, and the model interacts with the Interface using
discrete actions. This work examines the feasibility of learning to
interact with discrete Interfaces using the reinforce algorithm.


Discrete Interfaces cannot be trained directly with standard
backpropagation because they are not differentiable. It is most
natural to learn to interact with discrete Interfaces using
Reinforcement Learning methods. In this work, we consider an Input
Tape and a Memory Tape interface with discrete access.  Our concrete
proposal is to use the Reinforce algorithm  to learn
\emph{where} to access the discrete interfaces, and to use the
backpropagation algorithm to determine \emph{what} to write to the
memory and to the output.  We call this model the RL--NTM.

Discrete Interfaces are computationally attractive because the cost of
accessing a discrete Interface is often independent of its size.  It
is not the case for the continuous Interfaces, where the cost of
access scales linearly with size. It is a significant disadvantage
since slow models cannot scale to large difficult problems that require
intensive training on large datasets.  In
addition, an output Interface that lets the model decide when it wants to make
a prediction allows the model's runtime to be in principle unbounded.  If the
model has an output interface of this kind together with an interface to an unbounded
memory, the model becomes Turing complete.

We evaluate the RL-NTM on a number of simple algorithmic tasks.  The
RL-NTM succeeds on problems such as copying an input several times to
the output tape (the ``repeat copy'' task from
\cite{ntm}), reversing a sequence, and a few more tasks of
comparable difficulty. However, its success is highly dependent on the
architecture of the ``controller''.
We discuss this in more details in Section \ref{sec:flow}.

Finally, we found it non-trivial to correctly implement the RL-NTM due
its large number of interacting components. We developed a simple
procedure to numerically check the gradients of the Reinforce algorithm
(Section \ref{sec:checking}).  The procedure can be applied to
problems unrelated to NTMs, and is of the independent interest. The code for this work can be found at 
\href{https://github.com/ilyasu123/rlntm}{\color{blue} {https://github.com/ilyasu123/rlntm}}.

\vspace{-2mm}

\section{The Model}
\label{sec:model}

Many difficult tasks require a prolonged, multi-step interaction with
an external environment. Examples of such environments include computer games
\citep{mnih2013playing}, the stock market, an advertisement system, or
the physical world \citep{sergey}.  A model can observe a partial state
from the environment, and influence the environment through its
actions. This is seen as a general reinforcement leaning problem. 
However, our setting departs from the classical RL, i.e. 
we have a freedom to design tools available to solve a given problem. 
Tools might cooperate with the model (i.e. backpropagation
through memory), and the tools specify the actions over the environment. 
We formalize this concept under the name
Interface--Controller interaction.

The external environment is exposed to the model through a number
of Interfaces, each with its own API. For instance, a human perceives the
world through its senses, which include the vision Interface and the touch
Interface. The touch Interface provides methods for contracting the
various muscles, and methods for sensing the current state of the muscles,
pain level, temperature and a few others.  In this work, we explore a
number of simple Interfaces that allow the controller to access an
input tape, a memory tape, and an output tape.

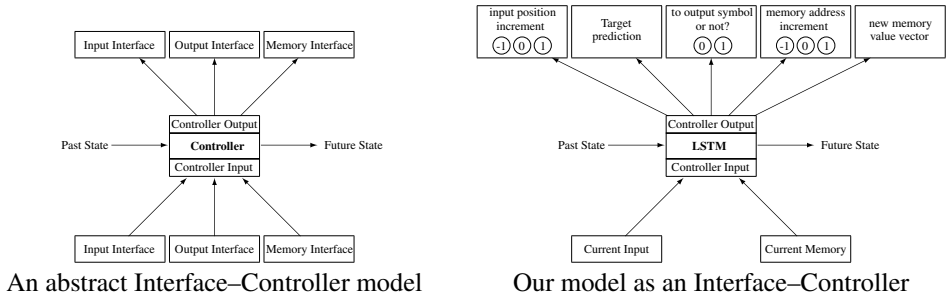
\begin{figure}[h]
    \begin{minipage}{0.45\linewidth}
      \centering
      \scalebox{0.65}{
        \begin{picture}(200, 160)
          \put(74, 60){\framebox(52, 15){}}
          \put(86, 65){{\scriptsize \textbf{Controller}}}
          \put(74, 50){\framebox(52, 10){}}
          \put(77, 53){{\scriptsize Controller Input}}
          \put(74, 75){\framebox(52, 10){}}
          \put(75, 78){{\scriptsize Controller Output}}

          \put(50, 15){\vector(1, 1){34}}
          \put(19, 0){\framebox(52, 15){}}
          \put(24, 5){{\scriptsize Input Interface}}

          \put(100, 15){\vector(0, 1){34}}
          \put(74, 0){\framebox(52, 15){}}
          \put(77, 5){{\scriptsize Output Interface}}

          \put(150, 15){\vector(-1, 1){34}}
          \put(129, 0){\framebox(52, 15){}}
          \put(130, 5){{\scriptsize Memory Interface}}

          \put(90, 85){\vector(-1, 1){34}}
          \put(19, 119){\framebox(52, 15){}}
          \put(24, 124){{\scriptsize Input Interface}}

          \put(100, 85){\vector(0, 1){34}}
          \put(74, 119){\framebox(52, 15){}}
          \put(77, 124){{\scriptsize Output Interface}}

          \put(115, 85){\vector(1, 1){34}}
          \put(129, 119){\framebox(52, 15){}}
          \put(130, 124){{\scriptsize Memory Interface}}

          \put(40, 68){\vector(1, 0){33}}
          \put(127, 68){\vector(1, 0){33}}
          \put(11, 66){{\scriptsize Past State}}
          \put(164, 66){{\scriptsize Future State}}

        \end{picture}
      }
    \\
    An abstract Interface--Controller model 
    \end{minipage}
    \begin{minipage}{0.45\linewidth}
    \centering
    \scalebox{0.65}{
      \begin{picture}(272, 160)
      \put(36, 0){
        \put(74, 60){\framebox(52, 15){}}
        \put(89, 65){{\scriptsize \textbf{LSTM}}}
        \put(74, 50){\framebox(52, 10){}}
        \put(77, 53){{\scriptsize Controller Input}}
        \put(74, 75){\framebox(52, 10){}}
        \put(75, 78){{\scriptsize Controller Output}}

        \put(50, 15){\vector(1, 1){34}}
        \put(19, 0){\framebox(52, 15){}}
        \put(26, 5){{\scriptsize Current Input}}

        \put(150, 15){\vector(-1, 1){34}}
        \put(129, 0){\framebox(52, 15){}}
        \put(131.5, 5){{\scriptsize Current Memory}}

        \put(75, 85){\vector(-2, 1){68}}
        \put(-36, 119){\framebox(52, 30){}}
        \put(-29, 143){{\scriptsize input position}}
        \put(-25, 135){{\scriptsize increment}}
        \put(-22, 126){\circle{10}}
        \put(-10, 126){\circle{10}}
        \put(2, 126){\circle{10}}
        \put(-25, 123.5){{\scriptsize -1}}
        \put(-12, 123.5){{\scriptsize 0}}
        \put(0.5, 123.5){{\scriptsize 1}}

        \put(90, 85){\vector(-1, 1){34}}
        \put(19, 119){\framebox(52, 30){}}
        \put(36, 137){{\scriptsize Target}}
        \put(31, 129){{\scriptsize prediction}}

        \put(100, 85){\vector(0, 1){34}}
        \put(74, 119){\framebox(52, 30){}}
        \put(77, 143){{\scriptsize to output symbol}}
        \put(91, 135){{\scriptsize or not?}}
        \put(95, 126){\circle{10}}
        \put(107, 126){\circle{10}}
        \put(93, 123.5){{\scriptsize 0}}
        \put(105.5, 123.5){{\scriptsize 1}}

        \put(110, 85){\vector(1, 1){34}}
        \put(129, 119){\framebox(52, 30){}}
        \put(132, 143){{\scriptsize memory address}}
        \put(140, 135){{\scriptsize increment}}
        \put(143, 126){\circle{10}}
        \put(155, 126){\circle{10}}
        \put(167, 126){\circle{10}}
        \put(140, 123.5){{\scriptsize -1}}
        \put(153, 123.5){{\scriptsize 0}}
        \put(165.5, 123.5){{\scriptsize 1}}

        \put(125, 85){\vector(2, 1){68}}
        \put(184, 119){\framebox(52, 30){}}
        \put(192, 137){{\scriptsize new memory}}
        \put(192.5, 129){{\scriptsize value vector}}

        \put(40, 68){\vector(1, 0){33}}
        \put(127, 68){\vector(1, 0){33}}
        \put(11, 66){{\scriptsize Past State}}
        \put(164, 66){{\scriptsize Future State}}
      }
      \end{picture}
    }
    \\
    Our model as an Interface--Controller
  \end{minipage} 

  \caption{\small \textbf{(Left)} The Interface--Controller
    abstraction, \textbf{(Right)} an instantiation of our model as an
    Interface--Controller. The bottom boxes are the read methods, and
    the top are the write methods. The RL--NTM makes discrete
    decisions regarding the move over the input tape, the memory tape,
    and whether to make a prediction at a given timestep. During
    training, the model's prediction is compared with the desired
    output, and is used to train the model when the RL-NTM chooses to
    advance its position on the output tape; otherwise it is
    ignored. The memory value vector is a vector of content that is
    stored in the memory cell.}
  \label{fig:interfaces}
\end{figure}

The part of the model that communicates with Interfaces is called the
Controller, which is the only part of the system which
learns. The Controller can have prior knowledge about behavior of
its Interfaces, but it is not the case in our experiments.  The
Controller learns to interact with Interfaces in a way that allows
it to solve a given task. Fig.~\ref{fig:interfaces} illustrates the complete
Interfaces--Controller abstraction.

\begin{figure}[h]
  \centering
  \includegraphics[width=0.9\linewidth]{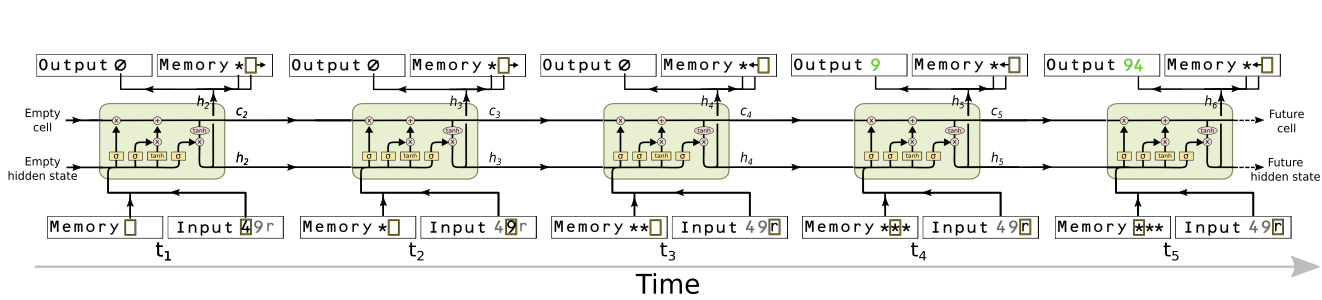}
  \caption{Execution of
    RL--NTM on the ForwardReverse task.  At each timestep, the RL-NTM
    consumes the value of the current input tape, the value of the
    current memory cell, and a representation of all the actions that
    have been taken in the previous timestep (not marked on the
    figures).  The RL-NTM then outputs a new value for the current
    memory cell (marked with a star), a prediction for the next target
    symbol, and discrete decisions for changing the positions of the heads 
    on the various tapes.  The RL-NTM learns to make discrete decisions using
    the Reinforce algorithm, and learns to produce continuous outputs
    using backpropagation.}
  \label{fig:exact}
\end{figure}

We now describe the RL--NTM.  As a controller, it uses either LSTM,
\emph{direct access}, or LSTM (see sec.~\ref{sec:modified-NTM} for a definition).
It has a one-dimensional input tape, a
one-dimensional memory, and a one-dimensional output tape as
Interfaces.  Both the input tape and the memory tape have a head that
reads the Tape's content at the current location.  The head of the input tape
and the memory tape can move in any direction.  However, the output
tape is a write-only tape, and its head can either stay at the current
position or move forward.  Fig.~\ref{fig:exact} shows an example
execution trace for the entire RL--NTM on the reverse task
(sec.~\ref{sec:tasks}).

At the core of the RL--NTM is an LSTM controller which receives
multiple inputs and has to generate multiple outputs at each timestep.
Table \ref{tab:input_output} summarizes the controller's inputs and
outputs, and the way in which the RL--NTM is trained to produce them.
The objective function of the RL--NTM is the expected log probability of
the desired outputs, where the expectation is taken over all possible
sequences of actions, weighted with probability of taking these
actions.  Both backpropagation and Reinforce maximize this objective.
Backpropagation maximizes the log probabilities of the model's predictions,
while the reinforce algorithm influences the probabilities of action
sequences.

The global objective can be written formally as:
\begin{align*}
  \sum_{[a_1, a_2, \dots, a_n] \in \mathbb{A}^\dagger} p_{\text{reinforce}}(a_1, a_2, \dots, a_n | \theta) \Big[\sum_{i=1}^n \log(p_{\text{bp}}(y_i | x_1, \dots, x_i, a_1, \dots a_i, \theta)\Big]
\end{align*}
$\mathbb{A}^\dagger$ represents the space of sequences of actions that
lead to the end of episode.  The probabilities in the above equation are
parametrized with a neural network (the Controller). We have marked
with $p_{\text{reinforce}}$ the part of the equation which is learned with
Reinforce.  $p_{\text{bp}}$ indicates the part of the equation
optimized with the classical backpropagation.

\begin{table}[h]
\centering
\renewcommand{\arraystretch}{1}
\begin{tabular}{|lcccc|}
\hline 
{\small Interface} &  & {\small Read} & {\small Write} & {\small Training Type} \\
\hline 
\tiny Input Tape & \tiny Head & \tiny window of values surrounding the current position & \tiny distribution over $[-1, 0, 1]$ & \tiny Reinforce \\
\hline
\multirow{ 2}{*}{\tiny Output Tape} & \tiny Head & $\varnothing$ & \tiny distribution over $[0, 1]$ & \tiny Reinforce \\ 
 & \tiny Content & $\varnothing$ & \tiny distribution over output vocabulary & \tiny Backpropagation \\ 
\hline
\multirow{ 2}{*}{\tiny Memory Tape} & \tiny Head & \multirow{ 2}{*}{\tiny window of memory values surrounding the current address} & \tiny distribution over $[-1, 0, 1]$  & \tiny Reinforce \\
& \tiny Content &  & \tiny vector of real values to store & \tiny Backpropagation \\
\hline
\tiny Miscellaneous & & \tiny all actions taken in the previous time step & $\varnothing$ & $\varnothing$ \\ 
\hline 
\end{tabular}
\caption{Table summarizes what the Controller reads at every time step, and what it has to produce.   
The ``training'' column indicates how the given part of the model is trained.}
\label{tab:input_output}
\end{table}

The RL--NTM receives a direct learning signal only when it decides to
make a prediction.  If it chooses to not make a prediction at a given
timestep, then it will not receive a direct learning signal.
Theoretically, we can allow the RL--NTM to run for an arbitrary number
of steps without making any prediction, hoping that after
sufficiently many steps, it would decide to make a prediction.  Doing
so will also provide the RL--NTM with arbitrary computational
capability.  However, this strategy is both unstable and
computationally infeasible.  Thus, we resort to limiting the total
number of computational steps to a fixed upper bound,
and force the RL--NTM to predict the next desired output whenever the
number of remaining desired outputs is equal to the number of
remaining computational steps.

\section{Related work}
\label{sec:related}

This work is the most similar to the Neural Turing Machine
\cite{ntm}.  The NTM is an ambitious, computationally universal model that
can be trained (or ``automatically programmed'') with the
backpropagation algorithm using only input-output examples.

Following the introduction NTM, several other memory-based models have
been introduced. All of them can be seen as part of a larger community
effort.  These models are constructed according to the
Interface--Controller abstraction (Section \ref{sec:model}).

\textbf{Neural Turing Machine} (NTM) \citep{graves2014neural} has a
modified LSTM as the Controller, and the following three Interfaces:
a sequential input, a delayed Output, and a differentiable Memory.

\vspace{-2mm} \textbf{Weakly supervised Memory Network}
\citep{sukhbaatar2015weakly} uses a feed forward network as the
Controller, and has a differentiable soft-attention Input, and Delayed
Output as Interfaces.

\vspace{-2mm} \textbf{Stack RNN} \citep{joulin2015inferring} has a
RNN as the Controller, and the sequential input, a differentiable
memory stack, and sequential output as Interfaces.  Also uses search
to improve its performance. 

\vspace{-2mm} \textbf{Neural DeQue} \citep{grefenstette2015learning}
has a LSTM as the Controller, and a Sequential Input, a differentiable
Memory Queue, and the Sequential Output as Interfaces.

\vspace{-2mm} \textbf{Our model} fits into the Interfaces--Controller
abstraction. It has a direct access LSTM as the Controller (or LSTM or
feed forward network), and its three interfaces are the Input Tape, the
Memory Tape, and the Output Tape.  All three Interfaces of the RL--NTM
are discrete and cannot be trained only with backpropagation.

This prior work investigates continuous and differentiable Interfaces,
while we consider discrete Interfaces.  Discrete Interfaces are more
challenging to train because backpropagation cannot be used.  However,
many external Interfaces are inherently discrete, even though humans
can easily use them (apparently without using continuous
backpropagation).  For instance, one interacts with the Google search
engine with discrete actions. This work examines the possibility of
learning models that interact with discrete Interfaces with the
Reinforce algorithm.

The Reinforce algorithm \citep{reinforce} is a classical RL algorithm,
which has been applied to the broad spectrum of planning problems
\citep{peters2006policy, kohl2004policy, aberdeen2002scaling}.  In
addition, it has been applied in object recognition to implement
visual attention \citep{mnih_attention, jimmy}.  This work uses
Reinforce to train an attention mechanism: we use it to train how to access
the various tapes provided to the model.

The RL--NTM can postpone prediction for an arbitrary number of
timesteps, and in principle has access to the unbounded memory. As a
result, the RL-NTM is Turing complete in principle.  There have been
very few prior models that are Turing complete
\cite{schmidhuber2012self, schmidhuber2004optimal}.  Although our
model is Turing complete, it is not very powerful because it is very
difficult to train, and our model can solve only relatively simple
problems.  Moreover, the RL--NTM does not exploit Turing completeness,
as none of tasks that it solves require superlinear runtime to be
solved.

\section{The Reinforce Algorithm}
\label{sec:Reinforce}
\textbf{Notation} \\
Let $\mathbb{A}$ be a space of actions, and $\mathbb{A}^\dagger$ be a space of all sequences of actions
that cause an episode to end (so $\mathbb{A}^\dagger \subset \mathbb{A}^*$) . An action at time-step $t$ is denoted by
$a_t$. We denote time at the end of episode by $T$ (this is not completely formal as some episodes can vary in time).
Let $a_{1:t}$ stand for a sequence of actions $[a_1, a_2, \dots, a_t]$.
Let $r(a_{1:t})$ denote the reward achieved at time $t$, having executed the sequence of actions $a_{1:t}$, 
and $R(a_{1:T})$ is the cumulative reward, namely $R(a_{k:T}) = \sum_{t=k}^T r(a_{1:t})$.
Let $p_\theta(a_t | a_{1:(t-1)})$ be a parametric conditional probability of an action $a_t$ given
all previous actions $a_{1:(t-1)}$. Finally, $p_\theta$ is a policy parametrized by $\theta$.

This work relies on learning discrete actions with the Reinforce
algorithm \citep{reinforce}.  We now describe this algorithm in
detail. Moreover, the supplementary materials include descriptions of
techniques for reducing variance of the gradient estimators.

The goal of reinforcement learning is to maximize the sum of future
rewards.  The Reinforce algorithm \citep{reinforce} does so directly by
optimizing the parameters of the policy $p_\theta(a_t | a_{1:(t -
  1)})$. Reinforce follows the gradient of the sum of the future
rewards.  The objective function for episodic reinforce can be expressed
as the sum over all sequences of valid actions that cause the episode to end:
\begin{align*}
  J(\theta) = \sum_{[a_1, a_2, \dots, a_T] \in\mathbb{A}^{\dagger}} p_\theta(a_1, a_2, \dots, a_T)R(a_1, a_2, \dots, a_T) = \sum_{a_{1:T} \in\mathbb{A}^{\dagger}} p_\theta(a_{1:T})R(a_{1:T}) 
\end{align*}
This sum iterates over sequences of all possible actions. This set 
is usually exponential or even infinite, 
so it cannot be computed exactly and cheaply for most of problems. However, 
it can be written as expectation, which can be approximated
with an unbiased estimator. We have that:
\begin{align*}
  J(\theta) = &\sum_{a_{1:T} \in\mathbb{A}^{\dagger}} p_\theta(a_{1:T})R(a_{1:T}) = \\
  &\mathbb{E}_{a_{1:T} \sim p_\theta} \sum_{t=1}^n r(a_{1:t}) = \\
  &\mathbb{E}_{a_1 \sim p_\theta(a_1)}
  \mathbb{E}_{a_2 \sim p_\theta(a_2 | a_1)} \dots
  \mathbb{E}_{a_T \sim p_\theta(a_T | a_{1:(T-1)})} 
   \sum_{t=1}^T r(a_{1:t}) 
\end{align*}
The last expression suggests a procedure to estimate $J(\theta)$:
simply sequentially sample each $a_t$ from the model distribution
$p_\theta(a_t | a_{1:(t - 1)})$ for $t$ from $1$ to $T$.
The unbiased estimator of $J(\theta)$ is the sum of $r(a_{1:t})$.
This gives us an algorithm to estimate $J(\theta)$.  However, the main
interest is in training a model to maximize this quantity.

The reinforce algorithm maximizes $J(\theta)$ by following the gradient of
it:
\begin{align*}
  \partial_{\theta} J(\theta) = \sum_{a_{1:T} \in \mathbb{A}^{\dagger}} \big[\partial_\theta p_\theta(a_{1:T})\big] R(a_{1:T})
\end{align*}
However, the above expression is a sum over the set of the possible
action sequences, so it cannot be computed directly for most
$\mathbb{A}^{\dagger}$. Once again, the Reinforce algorithm rewrites this sum as
an expectation that is approximated with sampling. It relies on the
equation: {${\partial_\theta f(\theta) = f(\theta) \frac{\partial_\theta f(\theta)}{f(\theta)} = f(\theta) \partial_\theta [\log f(\theta)]}$}.
This identity is valid as long as $f(x) \neq 0$. As typical neural
network parametrizations of distributions assign non-zero probability
to every action, this condition holds for $f = p_\theta$.  We have
that:
\begin{align*}
  \partial_\theta J(\theta) &= \sum_{[a_{1:T}] \in \mathbb{A}^{\dagger}} \big[\partial_\theta p_\theta(a_{1:T})\big] R(a_{1:T}) = \\
  &= \sum_{a_{1:T} \in \mathbb{A}^{\dagger}} p_\theta(a_{1:T}) \big[\partial_\theta \log p_\theta(a_{1:T})\big] R(a_{1:T}) \\
  &= \sum_{a_{1:T} \in \mathbb{A}^{\dagger}} p_\theta(a_{1:T}) \big[\sum_{t=1}^n \partial_\theta \log p_\theta(a_i | a_{1:(t - 1)})\big] R(a_{1:T})\\
  &= \mathbb{E}_{a_1 \sim p_\theta(a_1)}
  \mathbb{E}_{a_2 \sim p_\theta(a_2 | a_1)} \dots
  \mathbb{E}_{a_T \sim p_\theta(a_T | a_{1:{T-1}})} 
   \big[\sum_{t=1}^T \partial_\theta \log p_\theta(a_i | a_{1:(t - 1)}) \big]
   \big[\sum_{t=1}^T r(a_{1:t}) \big]
\end{align*}
The last expression gives us an algorithm for estimating $\partial_\theta J(\theta)$. 
We have sketched it at the left side of the Figure \ref{fig:reinforce}.
It's easiest to describe it with respect to computational graph behind a neural network. 
Reinforce can be implemented as follows. A neural network outputs: $l_t = \log p_\theta(a_t | a_{1:(t - 1)})$. 
Sequentially sample action $a_t$ from the distribution $e^{l_t}$, and execute the sampled action $a_t$.
Simultaneously, experience a reward $r(a_{1:t})$. Backpropagate the sum of the rewards 
$\sum_{t=1}^T r(a_{1:t})$ to the every node $\partial_\theta \log p_\theta(a_t | a_{1:(t - 1)})$.

We have derived an unbiased estimator for the sum of future rewards,
and the unbiased estimator of its gradient.  However, the derived
gradient estimator has high variance, which makes learning difficult.
RL--NTM employs several techniques to reduce gradient estimator
variance: (1) future rewards backpropagation, (2) online baseline
prediction, and (3) offline baseline prediction.  All these techniques
are crucial to solve our tasks. We provide detailed description of
techniques in the Supplementary material.

Finally, we needed a way of verifying the correctness of our implementation. 
We discovered a technique that makes it possible to easily implement a gradient checker
for nearly any model that uses Reinforce. Following Section \ref{sec:checking} describes 
this technique.

\section{Gradient Checking}
\label{sec:checking}
The RL--NTM is complex, so we needed to find an automated way of
verifying the correctness of our implementation.  We discovered a
technique that makes it possible to easily implement a gradient
checker for nearly any model that uses Reinforce. This discovery is an
independent contribution of this work.  This Section describes the
gradient checking for \emph{any} implementation of the reinforce
algorithm that uses a general function for sampling from multinomial
distribution.

\begin{figure}[h]
  \centering
  \fbox{
    \begin{minipage}[t]{0.46\linewidth}
      \centering
      \includegraphics[width=\linewidth]{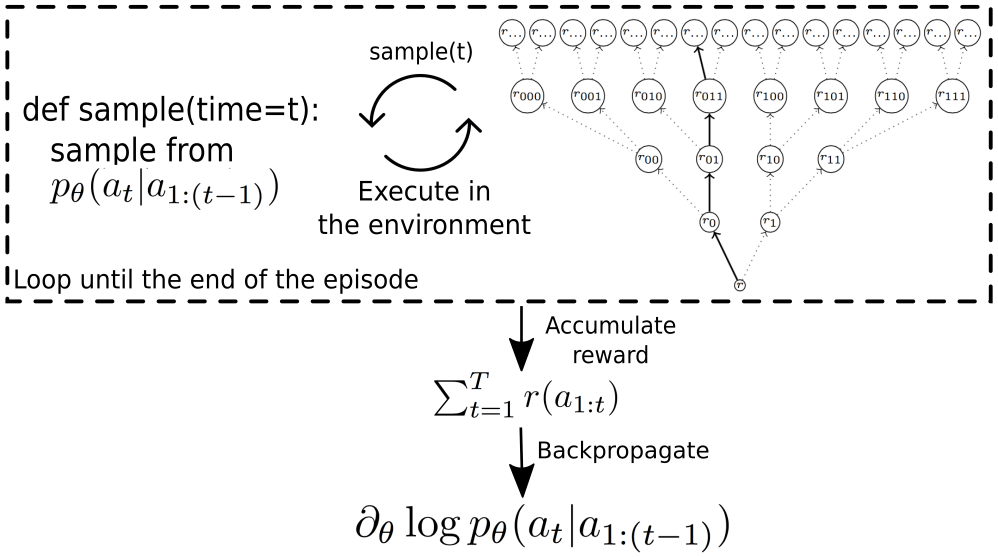}
      Reinforce
    \end{minipage}
  }
  \fbox{
    \begin{minipage}[t]{0.46\linewidth}
      \centering
      \includegraphics[width=\linewidth]{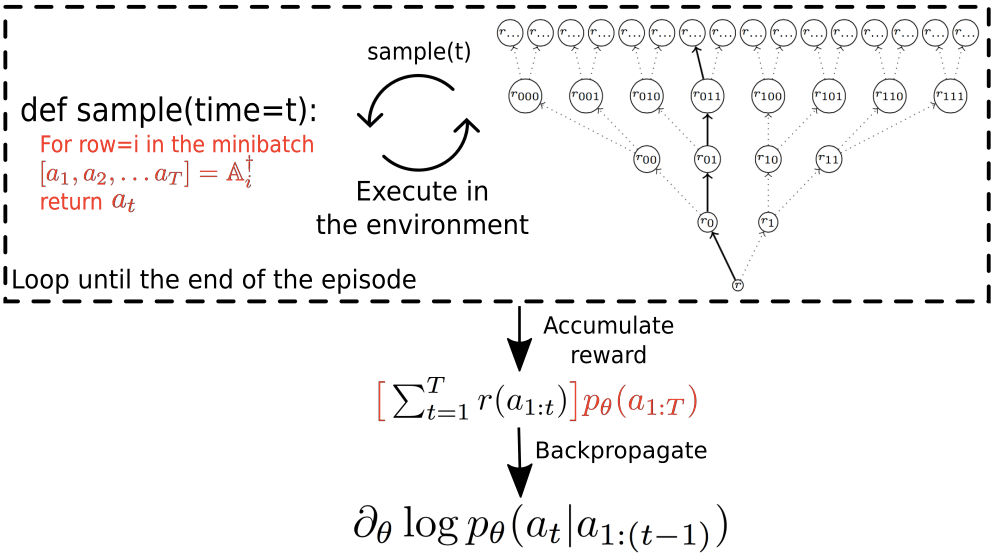}
      Gradient Checking of Reinforce
    \end{minipage}
  }
  \caption{Figure sketches algorithms: \textbf{(Left)} the reinforce algorithm, \textbf{(Right)} gradient checking for the reinforce algorithm. The red color indicates necessary steps to override the reinforce to become the gradient checker for the reinforce.}
  \label{fig:reinforce}
\end{figure}

The reinforce gradient verification should ensure that expected
gradient over all sequences of actions matches the numerical
derivative of the expected objective.  However, even for a tiny
problem, we would need to draw billions of samples to achieve
estimates accurate enough to state if there is match or mismatch.
Instead, we developed a technique which avoids sampling, and allows
for gradient verification of reinforce within seconds on a laptop.

First, we have to reduce the size of our a task to make sure that the
number of possible actions is manageable (e.g., $< 10^4$). This is
similar to conventional gradient checkers, which can only be applied
to small models.  Next, we enumerate all possible sequences
of actions that terminate the episode.  By definition, these are
precisely all the elements of $\mathbb{A}^\dagger$.  

The key idea is the following: we override the sampling function which
turns a multinomial distribution into a random sample with a
deterministic function that deterministically chooses actions from an
appropriate action sequence from $\mathbb{A}^\dagger$, while
accumulating their probabilities.  By calling the modified sampler, it
will produce every possible action sequence from $\mathbb{A}^\dagger$
exactly once.

For efficiency, it is desirable to use a single minibatch whose size
is $\#\mathbb{A}^\dagger$.  The sampling function needs to be adapted
in such a way, so that it incrementally outputs the appropriate
sequence from $\mathbb{A}^\dagger$ as we repeatedly call the sampling
function.  At the end of the minibatch, the sampling function will
have access to the total probability of each action sequence ($\prod_t
p_\theta(a_t|a_{1:t-1})$), which in turn can be used to exactly
compute $J(\theta)$ and its derivative.
To compute the derivative, the reinforce gradient produced 
by each  sequence $a_{1:T} \in \mathbb{A}^\dagger$ should be weighted by
its probability $p_\theta(a_{1:T})$.  We summarize this procedure on Figure
\ref{fig:reinforce}.

The gradient checking is critical for ensuring the correctness of our
implementation. While the basic reinforce algorithm is conceptually
simple, the RL--NTM is fairly complicated, as reinforce is used to
train several Interfaces of our model. Moreover, the RL--NTM uses
three separate techniques for reducing the variance of the gradient
estimators. The model's high complexity greatly increases the
probability of a code error.  In particular, our early implementations
were incorrect, and we were able to fix them only after implementing
gradient checking.

\section{Tasks}
\begin{figure}[h]
    \fbox{
    \begin{minipage}[t]{\linewidth}
      \centering
      \fbox{
        \begin{minipage}[t]{0.16\linewidth}
          \centering
          \vspace{2mm}
          \includegraphics[height=0.5\linewidth]{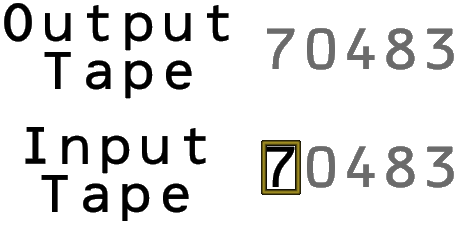}
          \vspace{2mm}
          {\small{\tiny\\}Copy}
          \vspace{3mm}
        \end{minipage}
      }
      \fbox{
        \begin{minipage}[t]{0.16\linewidth}
          \centering
          \vspace{2mm}
          \includegraphics[height=0.5\linewidth]{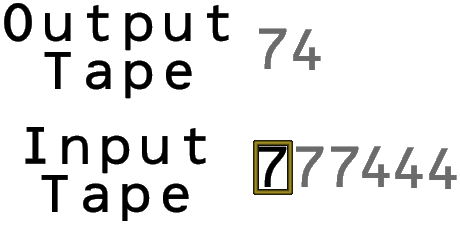}
          \vspace{2mm}
          {\small{\tiny\\}DuplicatedInput}
          \vspace{3mm}
        \end{minipage}
      }
      \fbox{
        \begin{minipage}[t]{0.16\linewidth}
          \vspace{2mm}
          \centering
          \includegraphics[height=0.5\linewidth]{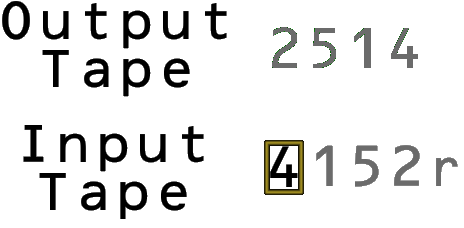}
          \vspace{2.5mm}
          {\small{\tiny\\}Reverse}
          \vspace{3mm}
        \end{minipage}
      }
      \fbox{
        \begin{minipage}[t]{0.16\linewidth}
          \vspace{2mm}
          \centering
          \includegraphics[height=0.5\linewidth]{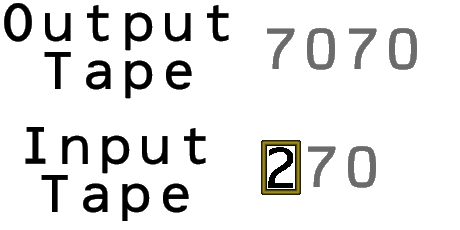}
          \vspace{2.5mm}
          {\small{\tiny\\}RepeatCopy}
          \vspace{2.5mm}
        \end{minipage}
      }
      \fbox{
        \begin{minipage}[t]{0.16\linewidth}
          \vspace{2mm}
          \centering
          \includegraphics[height=0.62\linewidth]{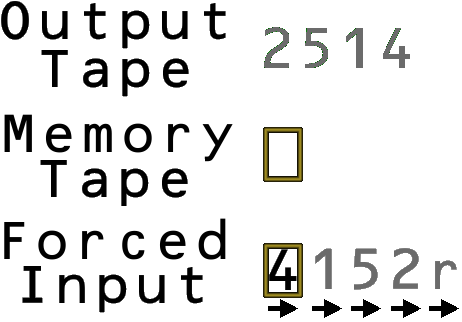}
          \vspace{2.5mm}
          {\small{\tiny\\}ForwardReverse}
          \vspace{3mm}
        \end{minipage}
      }
      \caption{This Figure presents the initial state for every task. The yellow box indicates the starting position
      of the reading head over the Input Interface.
      The gray characters on the Output Tape represent the target symbols.
      Our tasks involve reordering symbols, and 
      and the symbols $x_i$ have been picked uniformly from the set of size $30$.\\
      {\bf Copy.}  A generic input is $x_1 x_2 x_3 \ldots x_C\varnothing$ and the desired output is $x_1 x_2 \ldots x_C\varnothing$.  Thus the goal is to repeat the input. 
      The length of the input sequence is variable and is allowed to change.
      The input sequence and the desired output both terminate with a special end-of-sequence symbol $\varnothing$. \\
      {\bf DuplicatedInput.}  A generic input has the form $x_1 x_1 x_1 x_2 x_2 x_2 x_3 \ldots x_{C-1} x_C x_C x_C \varnothing$
      while the desired output is $x_1 x_2 x_3 \ldots x_C \varnothing$.  Thus each input symbol is replicated three times,
      so the RL-NTM must emit every third input symbol. \\
      {\bf Reverse.}  A generic input is  $x_1 x_2 \ldots x_{C-1} x_C\varnothing$ and the desired output is 
      $x_C x_{C-1} \ldots x_2 x_1\varnothing$. \\
      {\bf RepeatCopy.}  A generic input is $m x_1 x_2 x_3 \ldots x_C\varnothing$ and the desired output 
      is $x_1 x_2 \ldots x_C x_1 \ldots x_C x_1 \ldots x_C\varnothing$, where the number of copies is given by $m$.  Thus the goal is to copy
      the input $m$ times, where $m$ can be only 2 or 3. \\
      {\bf ForwardReverse.}  The task is identical to Reverse, but 
      the RL-NTM is only allowed to move its input tape pointer 
      forward. It means that a perfect solution must use the NTM's external memory.\\
  \label{fig:tasks}
      }
  \end{minipage}
  }

\end{figure}

\vspace{-2mm}
\label{sec:tasks}
This section defines tasks used in the experiments. 
Figure \ref{fig:tasks} shows exemplary instantiations of our tasks. Table
\ref{tab:available} summarizes the Interfaces that are available for each task.

\begin{table}[h]
\tiny
\centering
\renewcommand{\arraystretch}{1.15}
\begin{tabular}{|lcc|}
\hline 
\multicolumn{1}{|l|}{\backslashbox{Task}{Interface}} & Input Tape & Memory Tape \\
\hline 
Copy & \yes & \no \\
DuplicatedInput & \yes & \no \\ 
Reverse & \yes & \no \\
RepeatCopy & \yes & \no \\
ForwardReverse & \no & \yes \\ 
\hline 
\end{tabular}
\caption{This table marks the available Interfaces for each task.  The
  difficulty of a task is dependent on the type of Interfaces
  available to the model. }
\label{tab:available}
\end{table}

\section{Curriculum Learning}
\epigraph{Humans and animals learn much better when the examples are
  not randomly presented but organized in a meaningful order which
  illustrates gradually more concepts, and gradually more complex
  ones.  $\dots$ and call them ``curriculum learning''.}{\cite{bengio2009curriculum}}
We were unable to solve tasks with RL--NTM by training it on the
difficult instances of the problems (where difficult usually means
long).  To succeed, we had to create a curriculum of tasks of
increasing complexity.  We verified that our tasks were completely
unsolvable (in an all-or-nothing sense) for all but the shortest
sequences when we did not use a curriculum.  In our experiments, we
measure the complexity $c$ of a problem instance by the maximal length
of the desired output to typical inputs.  During training, we maintain
a distribution over the task complexity.  We shift the
distribution over the task complexities whenever the performance of
the RL--NTM exceeds a threshold.  Then, our model focuses on more
difficult problem instances as its performance improves.

\begin{table}[h]
\tiny
\centering
\renewcommand{\arraystretch}{1.15}
\begin{tabular}{|lc|}
\hline 
 Probability & Procedure to pick complexity $d$ \\
\hline 
$10\%$ & uniformly at random from the possible task complexities.  \\
$25\%$ & uniformly from $[1, C + e]$ \\ 
$65\%$ & $d = D + e$. \\
\hline 
\end{tabular}
\caption{The curriculum learning distribution, indexed by $C$. 
Here $e$ is a sample from a geometric distribution whose success probability is $\frac{1}{2}$, i.e., $p(e = k) = \frac{1}{2^k}$.}
\label{tab:procedure}
\end{table}

The distribution over task complexities is indexed with an integer
$c$, and is defined in Table \ref{tab:procedure}.  While we have not
tuned the coefficients in the curriculum learning setup, we
experimentally verified that it is critical to always maintain
non-negligible mass over the hardest difficulty levels
\citep{learning_to_execute}.  Removing it makes the curriculum much
less effective.

Whenever the average zero-one-loss (normalized by the length of the
target sequence) of our RL--NTM decreases below $0.2$, we increase
$c$ by 1. We kept doing so until $c$ reaches its maximal allowable
value.  Finally, we enforced a refractory period to ensure that
successive increments of $C$ are separated by at least $100$ parameter
updates, since we encountered situations where $C$ increased in rapid
succession which consistently caused learning to fail.

\section{Controllers}
\label{sec:flow}
The success of reinforcement learning training highly depends on
the complexity of the controller, and its ease of training.
It's common to either limit number of parameters of the
network, or to constraint it by initialization 
from pretrained model on some other task (for instance,
object recognition network for robotics).
Ideally, models should be generic enough to not
need such ``tricks''. However, still some 
tasks require building task specific architectures.

\begin{figure}[h]
\hspace{1cm}
\begin{minipage}{0.40\textwidth}
  \centering
  \includegraphics[width=\textwidth]{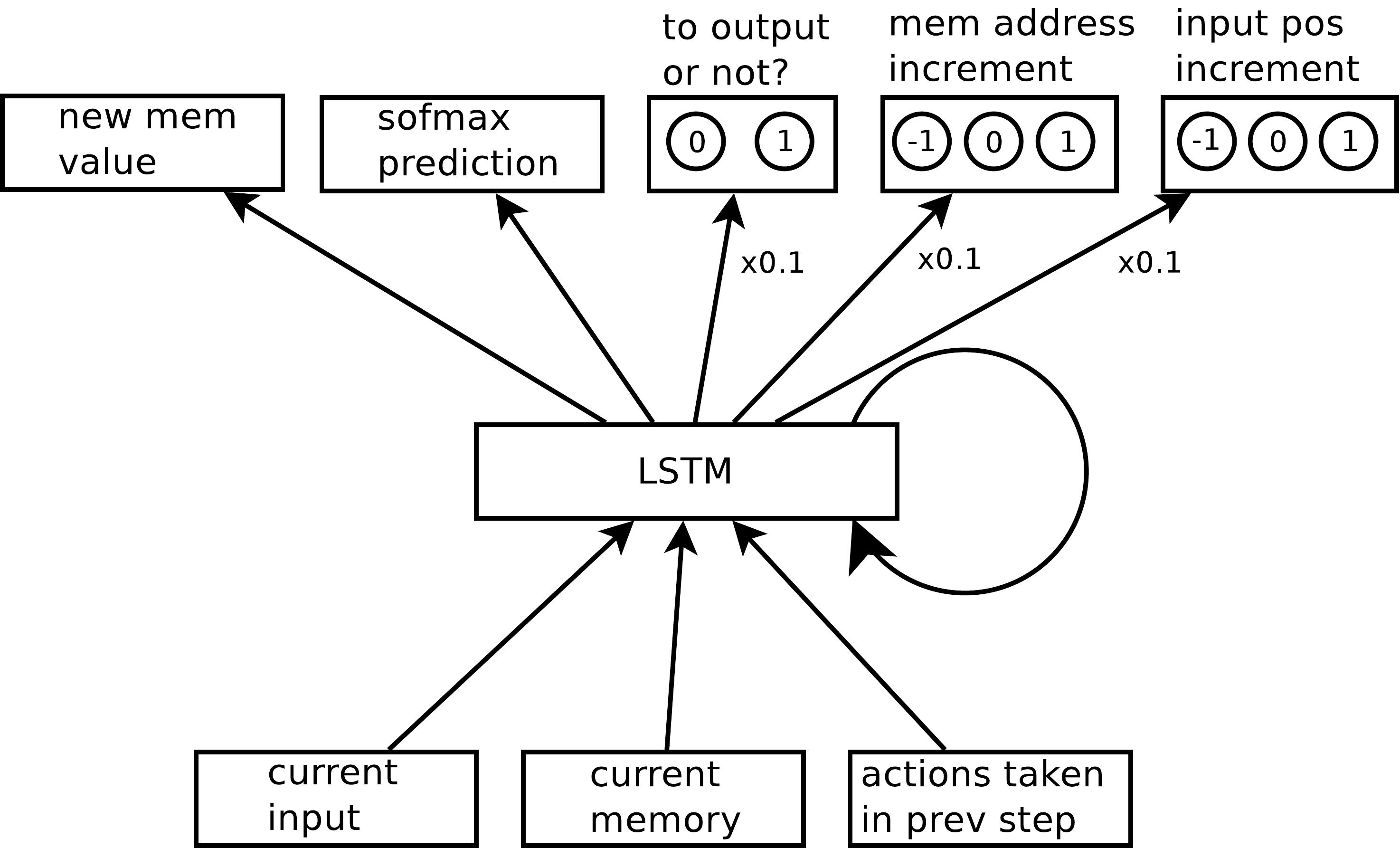} 
  \caption{\small \label{fig:NTM} LSTM as a controller. }
\end{minipage}
\hfill
\begin{minipage}{0.40\textwidth}
  \centering
  \includegraphics[width=\textwidth]{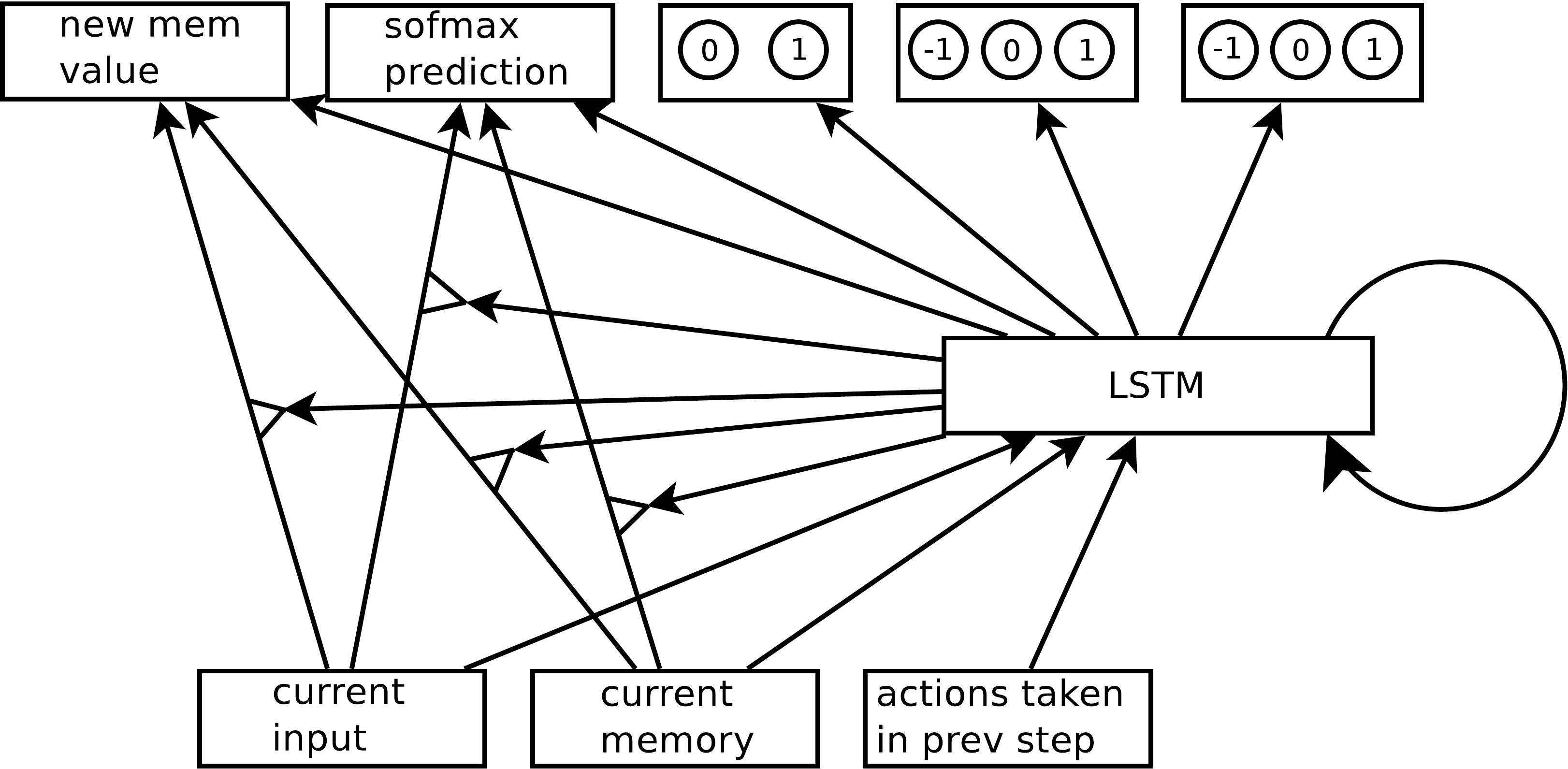} 
  \caption{\small \label{fig:modified-NTM} The direct access controller.}
\end{minipage}
\hspace{1cm}
\end{figure}

This work considers two controllers. The first is a LSTM (Fig.~\ref{fig:NTM}),
and the second is a direct access controller (Fig.~\ref{fig:modified-NTM}).
LSTM is a generic controller, that in principle should be powerful 
enough to solve any of the considered tasks. However, it
has trouble solving many of them. Direct access controller, is a much better fit for
symbol rearrangement tasks, however it's not a generic solution.

\subsection{Direct Access Controller}
\label{sec:modified-NTM}

All the tasks that we consider involve rearranging the input symbols
in some way.  For example, a typical task is to reverse a sequence
(section \ref{sec:tasks} lists the tasks).  For such tasks, the
controller would benefit from a built-in mechanism for directly
copying an appropriate input to memory and to the output.  Such a
mechanism would free the LSTM controller from remembering the input symbol in its control variables
(``registers''), and would shorten the backpropagation paths and
therefore make learning easier.  We implemented this mechanism by
adding the input to the memory and the output, and also adding the
memory to the output and to the adjacent memories (figure
\ref{fig:modified-NTM}), while modulating these additive contribution
by a dynamic scalar (sigmoid) which is computed from the controller's
state.  This way, the controller can decide to effectively not add the
current input to the output at a given timestep.  Unfortunately the
necessity of this architectural modification is a drawback of our
implementation, since it is not domain independent and would therefore
not improve the performance of the RL--NTM on many tasks of interest.

\begin{table}[h]
\tiny
\centering
\renewcommand{\arraystretch}{1.15}
\begin{tabular}{|lcc|}
\hline 
\multicolumn{1}{|l|}{\backslashbox{Task}{Controller}} & LSTM & Direct Access \\
\hline 
Copy & \yes & \yes \\
DuplicatedInput & \yes & \yes \\ 
Reverse & \no & \yes \\
ForwardReverse & \no & \yes \\ 
RepeatCopy & \no & \yes \\
\hline 
\end{tabular}
\caption{Success of training on various task for a given controller.}
\label{tab:controller}
\end{table}


\section{Experiments}
We presents results of training RL--NTM on all aforementioned tasks. 
The main drawback of our experiments is in the lack of comparison to the 
other models.
However, the tasks that we consider have to be considered in 
conjunction with available Interfaces, 
and other models haven't been considered with the same set of interfaces.
The statement, ``this model solves addition''
is difficult to assess, as the way that digits are delivered 
defines task difficulty.

The closest model to ours is NTM, and the shared task that they consider is copying.
We are able to generalize with copying to an arbitrary length. 
However, our Interfaces make this task very simple. Table \ref{tab:controller} 
summarizes results.

We trained our model using SGD with a fixed learning rate of $0.05$ and a fixed momentum of $0.9$. We used a batch of size $200$, 
which we found to work better than smaller batch sizes (such as $50$ or $20$).  We normalized the gradient by batch size but not by sequence length.
We independently clip the norm of the gradients w.r.t.~the RL-NTM parameters to $5$, and the gradient w.r.t.~the baseline network to $2$. 
We initialize the RL--NTM controller and the baseline model using a Gaussian with standard deviation $0.1$. 
We used an inverse temperature of $0.01$ for the different action distributions.  Doing so reduced the effective learning
rate of the Reinforce derivatives.
The memory consists of $35$ real values through which we backpropagate. 
The initial memory state and the controller's initial hidden states were set to the zero vector.

\begin{figure}[h]
\hspace{1cm}
\begin{minipage}{0.07\linewidth}
  \includegraphics[width=1.1\textwidth]{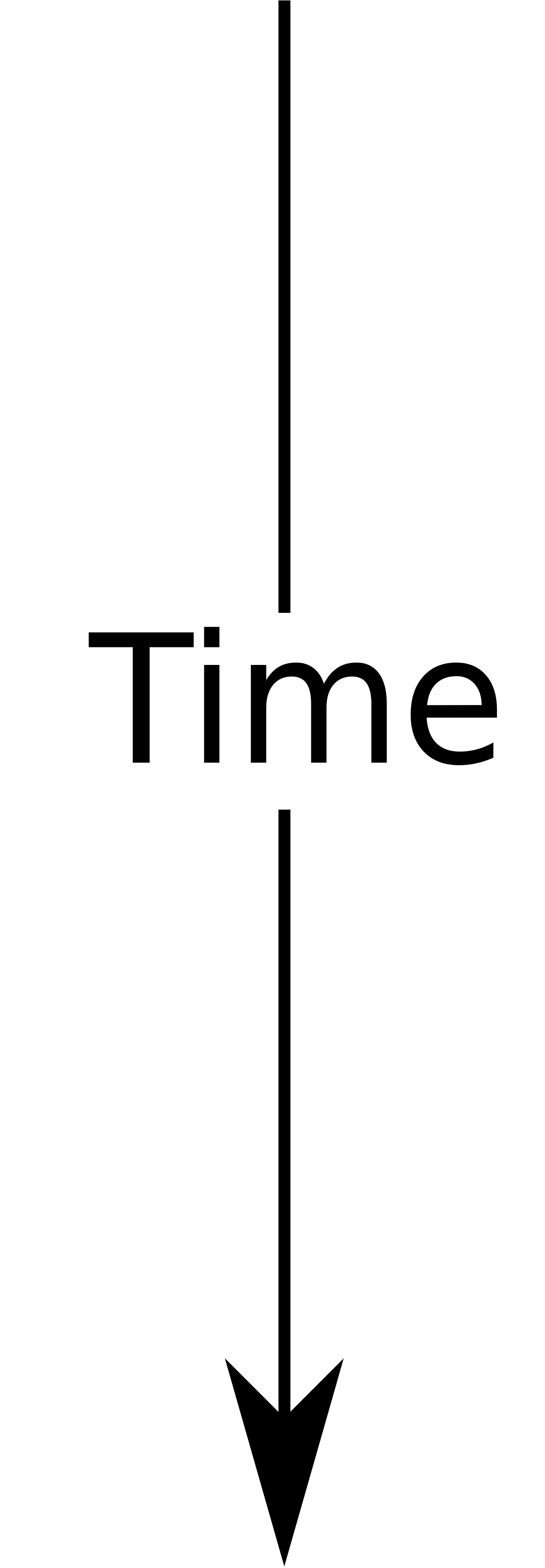} \\
\end{minipage}
\hspace{0.3cm}
\begin{minipage}{0.34\linewidth}
  \begin{center}
  \includegraphics[width=\textwidth]{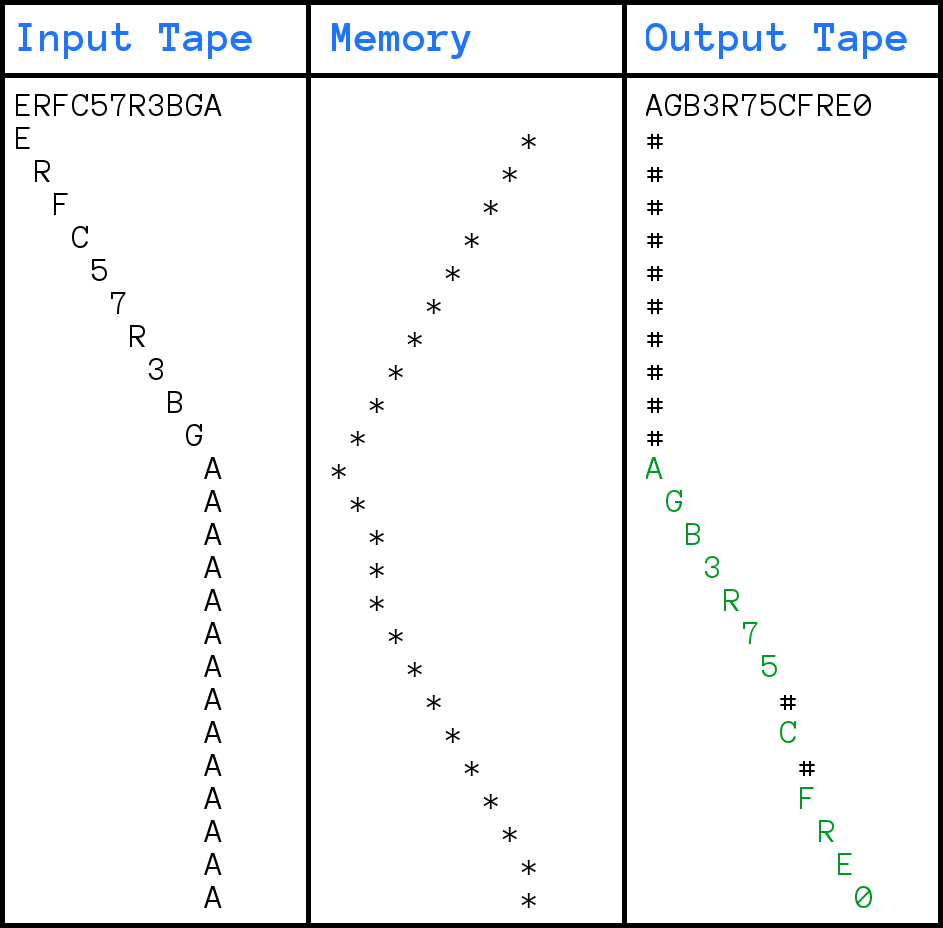} 
  \end{center}
\end{minipage}
\hfill
\begin{minipage}{0.30\linewidth}
  \begin{center}
    \includegraphics[width=\textwidth]{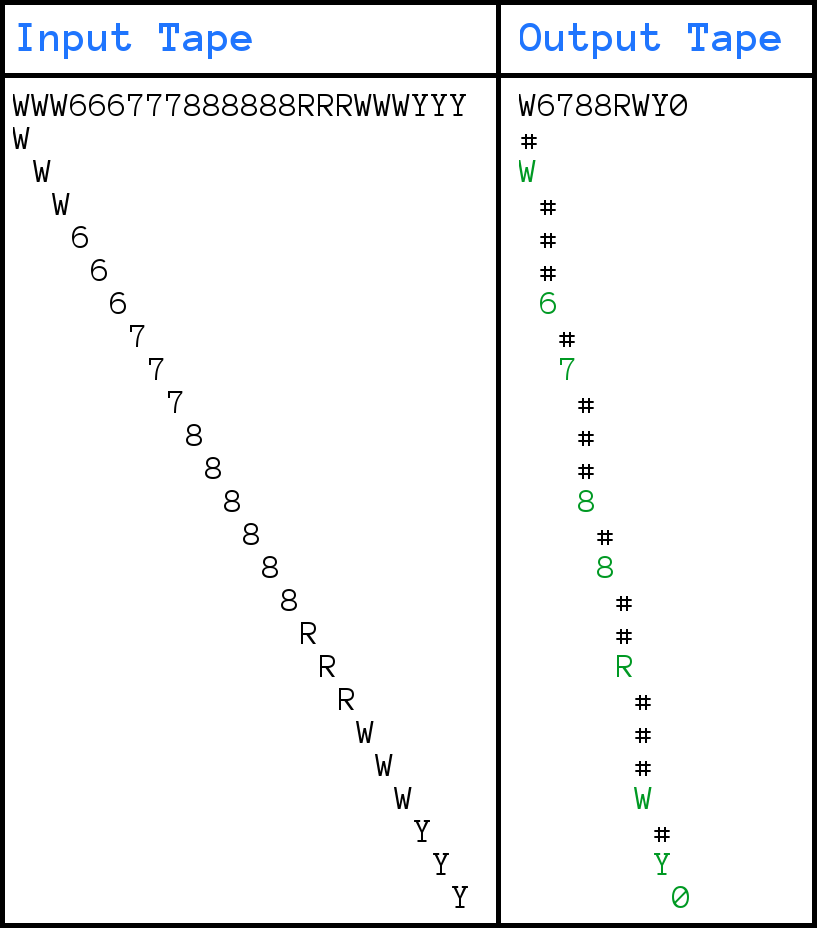} 
  \end{center}
\end{minipage}
\hspace{1cm}
  \caption{\textbf{(Left)} Trace of ForwardReverse solution,
           \textbf{(Right)} trace of RepeatInput.
    The vertical depicts execution time.  
  The rows show the input pointer, output pointer, 
  and memory pointer (with the $*$ symbol) at each step of the RL-NTM's execution.
  Note that we represent the set $\{1,\ldots,30\}$ with 30 distinct symbols, and lack of prediction 
  with $\#$.}
  \label{fig:traces}
\end{figure}

The ForwardReverse task is particularly interesting.  In order to solve the problem, the RL--NTM has to move
to the end of the sequence without making any predictions. 
While doing so, it has to store the input sequence into its memory (encoded in real values), 
and use its memory when reversing the sequence (Fig.~\ref{fig:traces}).

We have also experimented with a number of additional tasks but with less empirical success.  Tasks we found to be too difficult include 
sorting and long integer addition (in base 3 for simplicity), and RepeatCopy when the input tape is forced to only move forward.
While we were able to achieve reasonable performance on the sorting task, the RL--NTM learned an ad-hoc algorithm and made excessive use of its
controller memory in order to sort the sequence.

Empirically, we found all the components of the RL-NTM essential to successfully solving these problems.  
All our tasks are either solvable in under 20,000 parameter 
updates or fail in arbitrary number of updates. 
We were completely unable to solve RepeatCopy, Reverse, and Forward reverse with the LSTM controller, 
but with direct access controller we succeeded. 
Moreover, we were also unable to solve any of these problems at all without a curriculum (except for short sequences of length $5$).  
We present more traces for our tasks in the supplementary material (together
with failure traces).

\section{Conclusions} 

We have shown that the Reinforce algorithm is capable of training an NTM-style model
to solve very simple algorithmic problems.  While the Reinforce algorithm is very general and is 
easily applicable to a wide range of problems,
it seems that learning memory access patterns with Reinforce is difficult.

Our gradient checking procedure for Reinforce 
can be applied to a wide variety of implementations.  We also found it extremely useful: without it,
we had no way of being sure that our gradient was correct, which made debugging and tuning
much more difficult.

\section{Acknowledgments}
We thank Christopher Olah for the LSTM figure that have been used in
the paper, and to Tencia Lee for revising the paper.

{\small{
\bibliographystyle{iclr2016_conference}
\bibliography{rlntm} 
}
}

\newpage

\section*{Appendix A: Detailed Reinforce explanation}
We present here several techniques to decrease variance of the gradient estimation for 
the Reinforce. We have employed all of these tricks in our RL--NTM implementation.

We expand notation introduced in Sec.~\ref{sec:Reinforce}. 
Let $\mathcal{A}^\ddag$ denote all valid subsequences of actions (i.e. $\mathcal{A}^\ddag \subset \mathcal{A}^\dagger \subset \mathcal{A}^*$).
Moreover, we define set of sequences of actions that are valid after executing a sequence $a_{1:t}$, and that terminate. We denote such set by:
$\mathcal{A}_{a_{1:t}}^\dagger$. Every sequence $a_{(t+1):T} \in \mathcal{A}_{a_{1:t}}^\dagger$
terminates an episode.

\subsection*{Causality of actions}
Actions at time $t$ cannot possibly influence rewards obtained in the past, because
the past rewards are caused by actions prior to them.
This idea allows to derive an unbiased estimator of $\partial_\theta J(\theta)$ with
lower variance. Here, we formalize it:
\begin{align*}
  \partial_\theta J(\theta) &= \sum_{a_{1:T} \in \mathcal{A}^{\dagger}} p_\theta(a)\big[\partial_\theta \log p_\theta(a)\big] R(a)\\
  &= \sum_{a_{1:T} \in \mathcal{A}^{\dagger}} p_\theta(a)\big[\partial_\theta \log p_\theta(a)\big] \big[\sum_{t=1}^T r(a_{1:t})\big]\\
  &= \sum_{a_{1:T} \in \mathcal{A}^{\dagger}} p_\theta(a)\big[\sum_{t=1}^T \partial_\theta \log p_\theta(a_{1:t}) r(a_{1:t}) \big]\\
  &= \sum_{a_{1:T} \in \mathcal{A}^{\dagger}} p_\theta(a)\big[\sum_{t=1}^T \partial_\theta \log p_\theta(a_{1:t}) r(a_{1:t}) + \partial_\theta \log p_\theta({a_{(t+1):T}} | a_{1:t}) r(a_{1:t}) \big]\\
  &= \sum_{a_{1:T} \in \mathcal{A}^{\dagger}} \sum_{t=1}^T p_\theta(a_{1:t}) \partial_\theta \log p_\theta(a_{1:t}) r(a_{1:t}) + p_\theta(a) \partial_\theta \log p_\theta({a_{(t+1):T}} | a_{1:t}) r(a_{1:t}) \\
  &= \sum_{a_{1:T} \in \mathcal{A}^{\dagger}} \sum_{t=1}^T p_\theta(a_{1:t}) \partial_\theta \log p_\theta(a_{1:t}) r(a_{1:t}) + p_\theta(a_{1:t}) r(a_{1:t}) \partial_\theta p_\theta({a_{(t+1):T}} | a_{1:t}) \\
  &= \sum_{a_{1:T} \in \mathcal{A}^{\dagger}} \big[ \sum_{t=1}^T p_\theta(a_{1:t}) \partial_\theta \log p_\theta(a_{1:t}) r(a_{1:t}) \big] + \sum_{a_{1:T} \in \mathcal{A}^{\dagger}}  \sum_{t=1}^T \big[p_\theta(a_{1:t}) r(a_{1:t}) \partial_\theta p_\theta({a_{(t+1):T}} | a_{1:t}) \big] \\
\end{align*}

We will show that the right side of this equation is equal to zero. It's zero, because the future actions $a_{(t+1):T}$
don't influence past rewards $r(a_{1:t})$. Here we formalize it;
we use an identity $\mathbb{E}_{a_{(t+1):T} \in \mathcal{A}_{a_{1:t}}^\dagger} p_\theta(a_{(t+1):T} | a_{1:t}) = 1$:
\begin{align*}
  \sum_{a_{1:T} \in \mathcal{A}^{\dagger}}&  \sum_{t=1}^T \big[p_\theta(a_{1:t}) r(a_{1:t}) \partial_\theta p_\theta({a_{(t+1):T}} | a_{1:t}) \big] = \\
  \sum_{a_{1:t} \in \mathcal{A}^{\ddag}}& \big[p_\theta(a_{1:t}) r(a_{1:t}) \sum_{a_{(t+1):T} \in \mathcal{A}_{a_{1:t}}^{\dagger}} \partial_\theta p_\theta(a_{(t+1):T} | a_{1:t}) \big] = \\
  \sum_{a_{1:t} \in \mathcal{A}^{\ddag}}& p_\theta(a_{1:t}) r(a_{1:t}) \partial_\theta 1 = 0
\end{align*}
We can purge the right side of the equation for $\partial_\theta J(\theta)$:
\begin{align*}
  \partial_\theta J(\theta)
  &= \sum_{a_{1:T} \in \mathcal{A}^{\dagger}} \big[ \sum_{t=1}^T p_\theta(a_{1:t}) \partial_\theta \log p_\theta(a_{1:t}) r(a_{1:t}) \big] \\
  &= \mathbb{E}_{a_1 \sim p_\theta(a)}
  \mathbb{E}_{a_2 \sim p_\theta(a | a_1)} \dots
  \mathbb{E}_{a_T \sim p_\theta(a | a_{1:(T-1)})} 
  \big[\sum_{t=1}^T \partial_\theta \log p_\theta(a_t | a_{1:(t - 1)}) \sum_{i=\mathbf{t}}^T r(a_{1:i}) \big]
\end{align*}

The last line of derived equations describes the learning algorithm. 
This can be implemented as follows. A neural network outputs: $l_t = \log p_\theta(a_t | a_{1:(t - 1)})$. 
We sequentially sample action $a_t$ from the distribution $e^{l_t}$, and execute the sampled action $a_t$.
Simultaneously, we experience a reward $r(a_{1:t})$. We should backpropagate 
to the node $\partial_\theta \log p_\theta(a_t | a_{1:(t - 1)})$
the sum of rewards starting from time step $t$:  $\sum_{i=t}^T r(a_{1:i})$. The only difference
in comparison to the initial algorithm is that we backpropagate sum of rewards starting from the 
current time step, instead of the sum of rewards over the entire episode.

\subsection*{Online baseline prediction}
Online baseline prediction is an idea, that the importance of reward is determined by its relative relation 
to other rewards. All the rewards could be shifted by a constant factor and such change shouldn't 
effect its relation, thus it shouldn't influence expected gradient. However, it could decrease the variance of the gradient estimate.

Aforementioned shift is called the baseline, and it can be
estimated separately for the every time-step.
We have that:
\begin{align*}
  \sum_{a_{(t+1):T} \in \mathcal{A}_{a_{1:t}}^\dagger} p_\theta(a_{(t+1):T} | a_{1:t}) = 1 \\
  \partial_{\theta} \sum_{a_{(t + 1):T} \in \mathcal{A}_{a_{1:t}}^\dagger} p_\theta(a_{(t+1):T} | a_{1:t}) = 0 \\
\end{align*}
We are allowed to subtract above quantity (multiplied by $b_t$) from our estimate of the gradient without changing 
its expected value:
\begin{align*}
  \partial_\theta J(\theta)  
  = \mathbb{E}_{a_1 \sim p_\theta(a)} 
  \mathbb{E}_{a_2 \sim p_\theta(a | a_1)} \dots
  \mathbb{E}_{a_T \sim p_\theta(a | a_{1:(T-1)})} 
  \big[\sum_{t=1}^T \partial_\theta \log p_\theta(a_t | a_{1:(t - 1)}) \sum_{i=\mathbf{t}}^T (r(a_{1:i}) - b_t) \big]
\end{align*}
Above statement holds for an any sequence of $b_t$. 
We aim to find the sequence $b_t$ that yields the lowest variance estimator on $\partial_\theta J(\theta)$. 
The variance of our estimator is:
\begin{align*}
  Var = \mathbb{E}_{a_1 \sim p_\theta(a)}
  \mathbb{E}_{a_2 \sim p_\theta(a | a_1)} \dots
  \mathbb{E}_{a_T \sim p_\theta(a | a_{1:(T-1)})} 
  \big[\sum_{t=1}^T \partial_\theta \log p_\theta(a_t | a_{1:(t-1)}) \sum_{i=\mathbf{t}}^T (r(a_{1:i}) - b_t) \big]^2 - \\
  \Big[\mathbb{E}_{a_1 \sim p_\theta(a)}
  \mathbb{E}_{a_2 \sim p_\theta(a | a_1)} \dots
  \mathbb{E}_{a_T \sim p_\theta(a | a_{1:(T-1)})} 
  \big[\sum_{t=1}^T \partial_\theta \log p_\theta(a_t | a_{1:(t - 1)}) \sum_{i=\mathbf{t}}^T (r(a_{1:i}) - b_t) \big]\Big]^2
\end{align*}
The second term doesn't depend on $b_t$, and the variance is always positive. It's sufficient to minimize the first term. 
The first term is minimal when it's derivative with respect to $b_t$ is zero. This implies
\begin{align*}
  &\mathbb{E}_{a_1 \sim p_\theta(a)}
  \mathbb{E}_{a_2 \sim p_\theta(a | a_1)} \dots
  \mathbb{E}_{a_T \sim p_\theta(a | a_{1:(T-1)})} 
  \sum_{t=1}^T \partial_\theta \log p_\theta(a_t | a_{1:(t - 1)}) \sum_{i=\mathbf{t}}^T (r(a_{1:i}) - b_t) = 0 \\
  &\sum_{t=1}^T \partial_\theta \log p_\theta(a_t | a_{1:(t - 1)}) \sum_{i=\mathbf{t}}^T (r(a_{1:i}) - b_t) = 0 \\
  &b_t = \frac{\sum_{t=1}^T \partial_\theta \log p_\theta(a_t | a_{1:(t - 1)}) \sum_{i=\mathbf{t}}^T r(a_{1:t})}{\sum_{t=1}^T \partial_\theta \log p_\theta(a_t | a_{1:(t - 1)})}
\end{align*}
This gives us estimate for a vector $b_t \in \mathbb{R}^{\#\theta}$.
However, it is common to use a single scalar for $b_t \in \mathbb{R}$, and estimate it as
$\mathbb E_{p_\theta(a_{t:T} | a_{1:(t - 1)})} R(a_{t:T})$.

\subsection*{Offline baseline prediction}
\begin{figure}
\centerline{
\includegraphics[width=.8\textwidth]{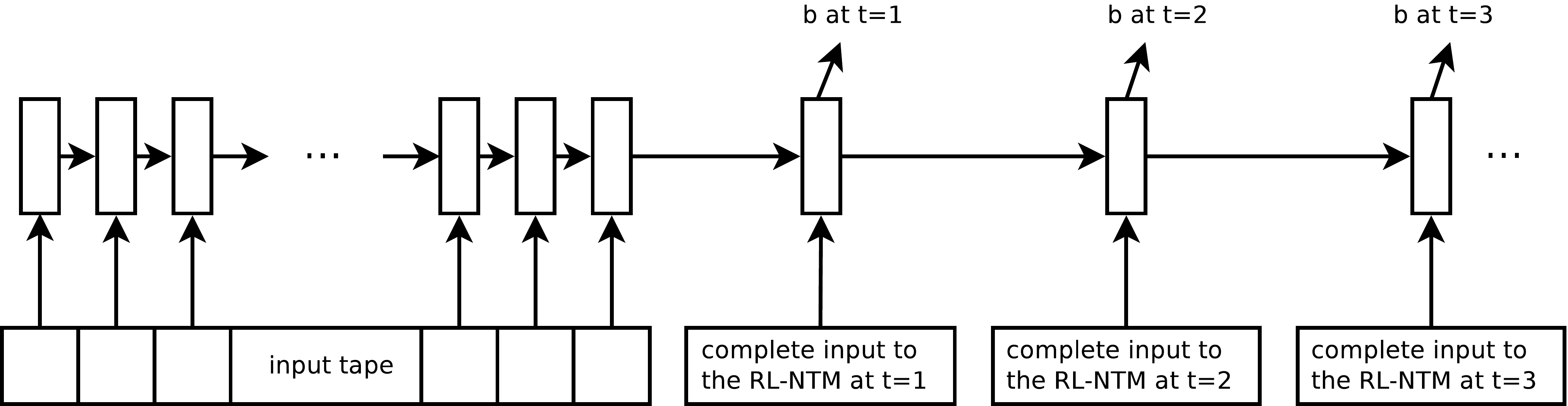}
}
\caption{The baseline LSTM computes a baseline $b_t$ for every computational step $t$ of the RL-NTM.  
The baseline LSTM receives the same inputs as the RL-NTM, and it computes a baseline $b_t$ for time $t$ before 
observing the chosen actions of time $t$.  However, it is important to first provide the baseline LSTM with the entire input tape
as a preliminary inputs, because doing so allows 
the baseline LSTM to accurately estimate the true difficulty of a given problem instance and therefore
compute better baselines.  For example, if a problem instance is unusually difficult, then we expect $R_1$ to be large and negative.  
If the baseline LSTM is given entire input tape as an auxiliary input, it could compute an appropriately large and negative $b_1$.}
\label{fig:baseline}
\end{figure}

The Reinforce algorithm works much better whenever it has accurate baselines.
A separate LSTM can help in the baseline estimation.
First, run the baseline LSTM on the entire input tape to produce a vector summarizing the input.
Next, continue running the baseline LSTM in tandem with the controller LSTM, so that the baseline LSTM receives
precisely the same inputs as the controller LSTM, and outputs a baseline $b_t$ at each timestep $t$.
The baseline LSTM is trained to minimize $\sum_{t=1}^T \big[ R(a_{t:T}) - b_t \big]^2$ (Fig.~\ref{fig:baseline}). This technique
introduces a biased estimator, however it works well in practise.

We found it important to first have the baseline LSTM go over the entire input  before computing 
the baselines $b_t$.  It is especially beneficial whenever there is considerable variation in the difficulty
of the examples.  For example, if the baseline LSTM can recognize that the current instance is unusually
difficult, it can output a large negative value for $b_{t=1}$ in anticipation of a large and a negative $R_1$.
In general, it is cheap and therefore worthwhile to provide the baseline network with all of the available
information, even if this information would not be available at test time, because the baseline network is not 
needed at test time.  

\section*{Appendix B: Execution Traces}
We present several execution traces of the RL--NTM.   
Each figure shows execution traces
of the trained RL-NTM on each of the tasks.  The first row shows the input tape
and the desired output, while each subsequent row shows the RL-NTM's position on the input
tape and its prediction for the output tape.  In these examples,  the RL-NTM solved each
task perfectly, so the predictions made in the output tape perfectly match the  
desired outputs listed in the first row. 

\begin{figure}[h]
  \begin{minipage}{0.48\textwidth}
    \begin{center}
      \includegraphics[width=0.7\textwidth]{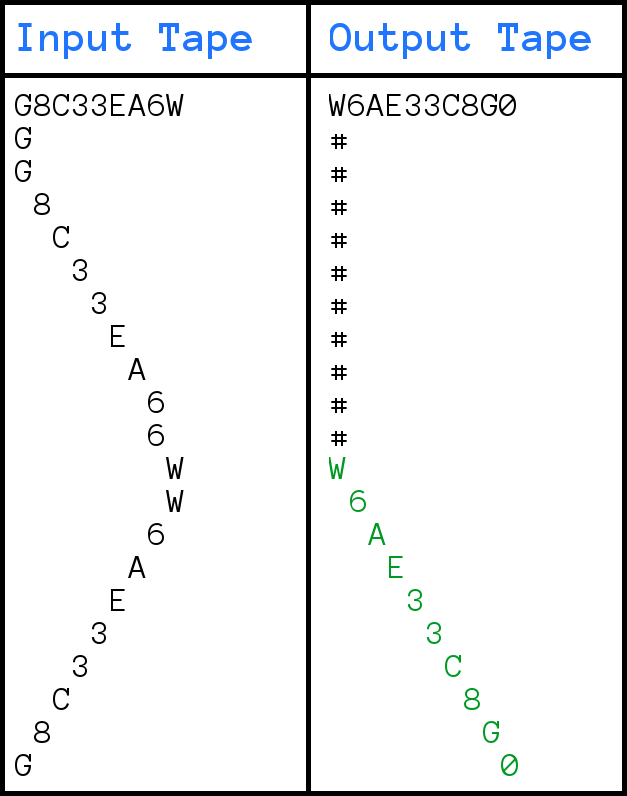} \\
    \end{center}
    An RL-NTM successfully solving a small instance of the Reverse problem
    (where the external memory is not used).
  \end{minipage}
  \hfill
  \begin{minipage}{0.48\textwidth}
    \includegraphics[width=\textwidth]{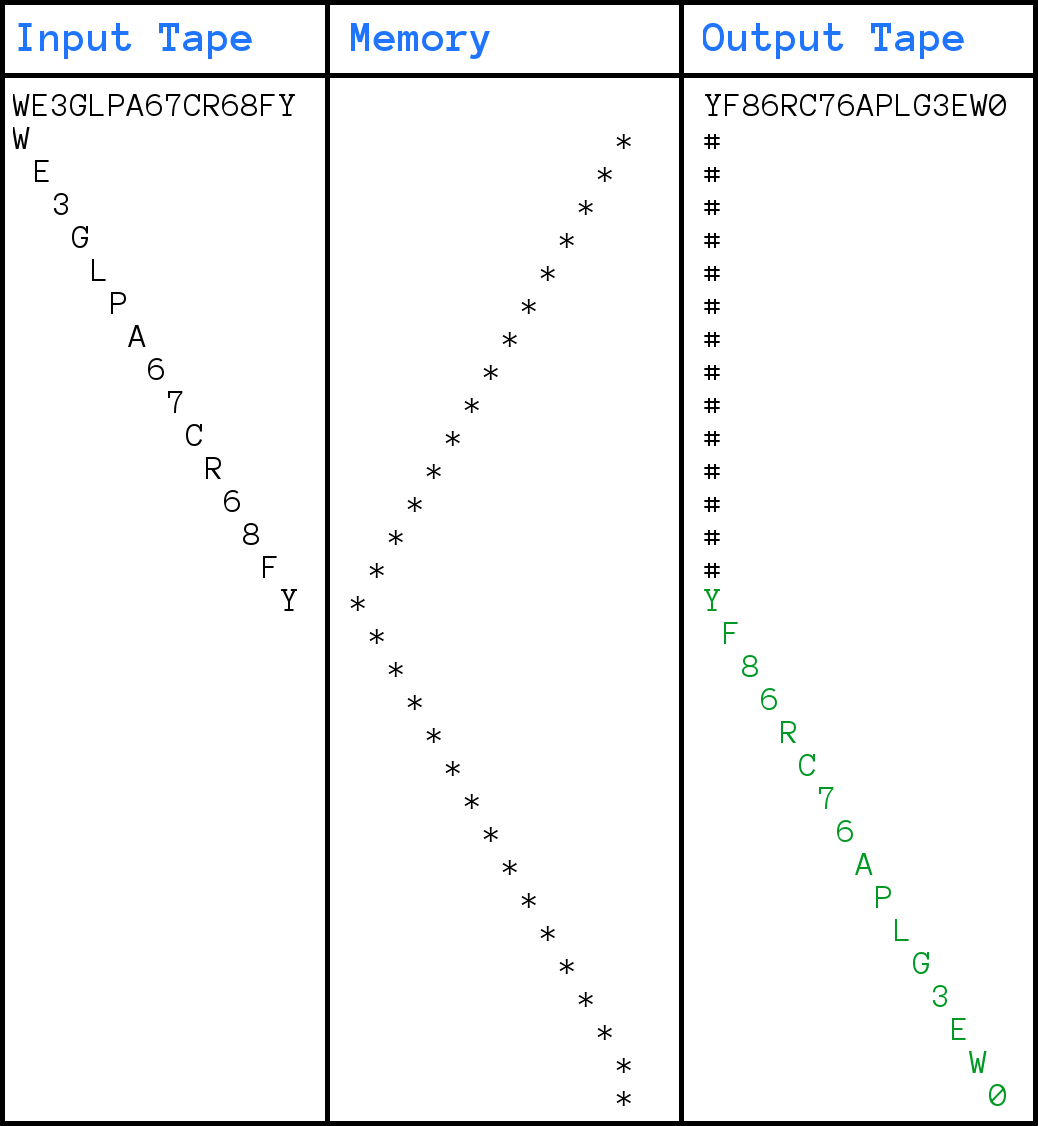} \\
    An RL-NTM successfully solving a small instance of the ForwardReverse problem,
    where the external memory is used. 
  \end{minipage}
\end{figure}

\begin{figure}[h]
  \begin{minipage}{0.38\textwidth}
    \begin{center}
      \includegraphics[width=0.8\textwidth]{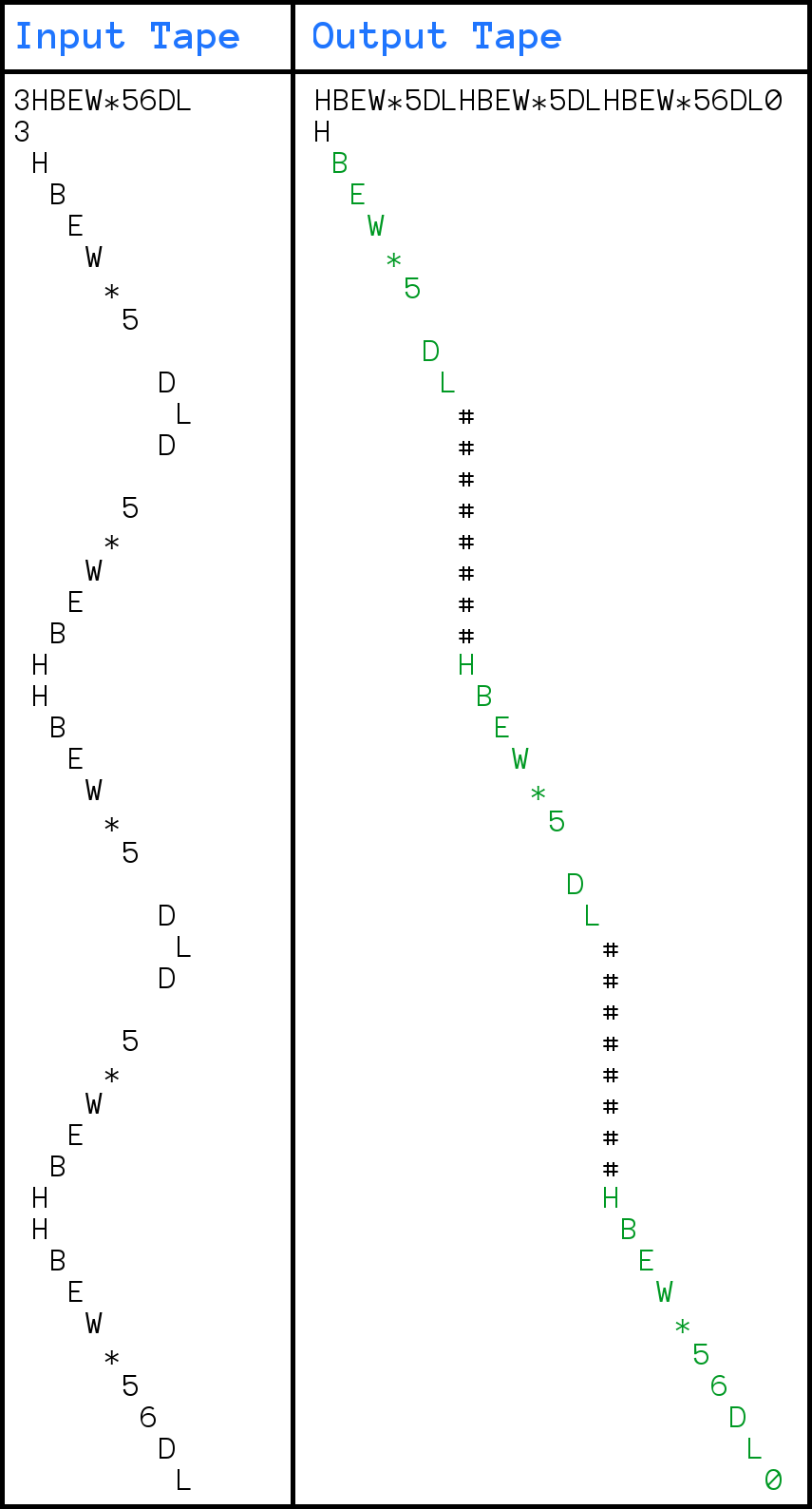} \\
    \end{center}
An RL-NTM successfully solving an instance of the RepeatCopy problem where the input
is to be repeated three times.
  \end{minipage}
  \hfill
  \begin{minipage}{0.58\textwidth}
    \begin{center}
    \includegraphics[width=0.7\textwidth]{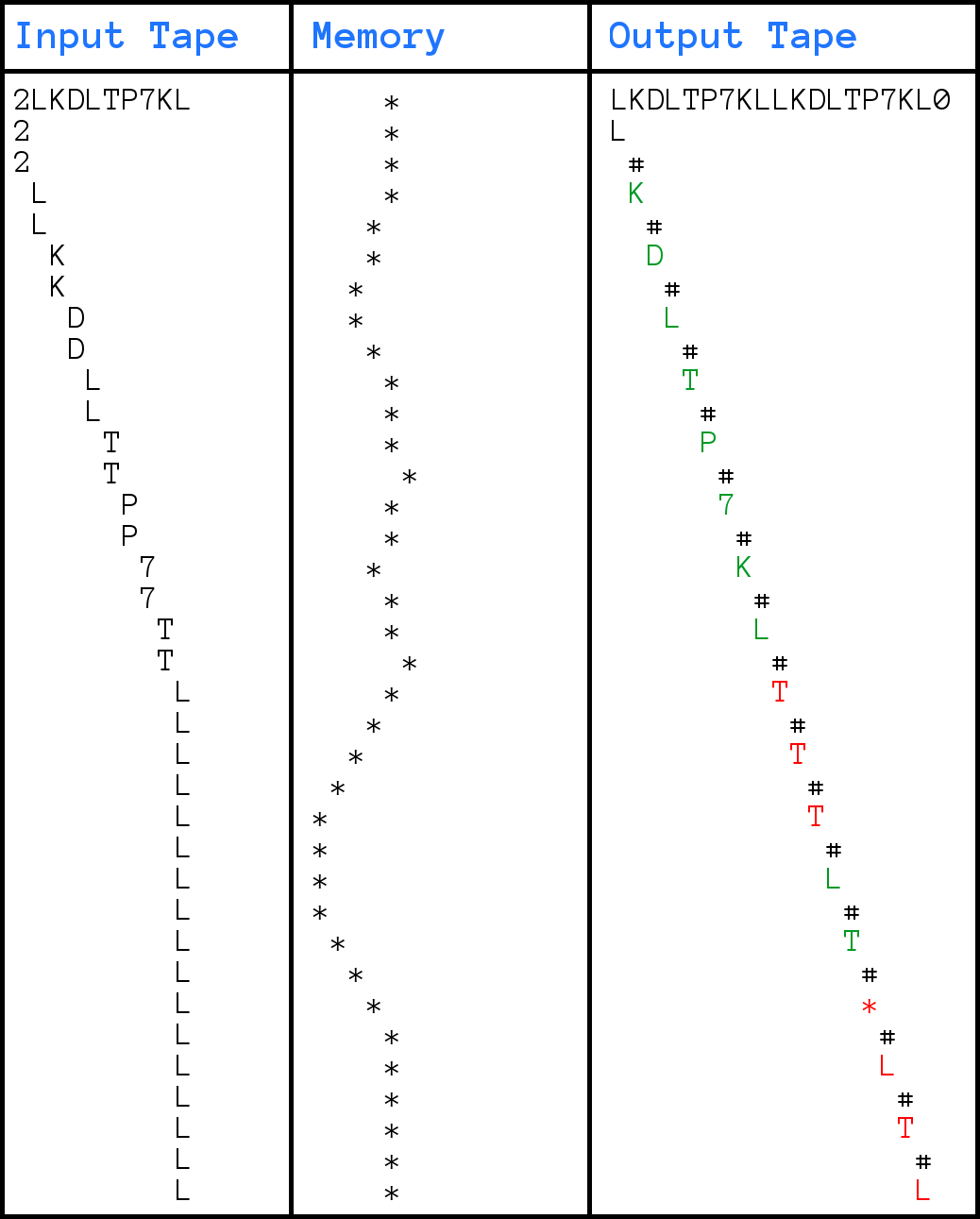} \\
    \end{center}
    An example of a failure of the RepeatCopy task, where the input tape is only allowed to move forward.
    The correct solution would have been to copy the input to the memory, and then solve the task using the memory.
    Instead, the memory pointer is moving randomly.
  \end{minipage}
\end{figure}

\end{document}